\documentclass[lettersize,journal]{IEEEtran}
\usepackage{amsmath,amsfonts}
\usepackage{algorithmic}
\usepackage{algorithm}
\usepackage{array}
\usepackage[caption=false,font=normalsize,labelfont=sf,textfont=sf]{subfig}
\usepackage{textcomp}
\usepackage{stfloats}
\usepackage{url}
\usepackage{verbatim}
\usepackage{graphicx}
\usepackage{cite}
\usepackage{xcolor}
\usepackage{multirow}
\usepackage{amsmath}

\begin{document}

\title{Generalized Few-Shot 3D Object Detection of LiDAR Point Cloud for Autonomous Driving}
\author{Jiawei Liu, Xingping~Dong, Sanyuan~Zhao and Jianbing~Shen,~\IEEEmembership{Senior Member,~IEEE},

\IEEEcompsocitemizethanks{

\IEEEcompsocthanksitem J. Liu is with the School of Computer Science \& Technology, Beijing Institute of Technology, China.
(email: jiawei\_liu@bit.edu.cn)
\IEEEcompsocthanksitem X. Dong is with the Inception Institute of Artificial Intelligence, Abu Dhabi, UAE.
(email: xingping.dong@gmail.com)
\IEEEcompsocthanksitem  S. Zhao is with the School of Computer Science \& Technology, Beijing Institute of Technology, China, and also with Yangtze Delta Region Academy of Beijing Institute of Technology, Jiaxing, China.
(email: zhaosanyuan@bit.edu.cn)
\IEEEcompsocthanksitem J. Shen is with the State Key Laboratory of Internet of Things for Smart City,
Department of Computer and Information Science, University of Macau, Macau, China.
(email: shenjianbingcg@gmail.com)

}
\thanks{}
}


\markboth{Journal of \LaTeX\ Class Files,~Vol.~14, No.~8, August~2021}%
{Shell \MakeLowercase{\textit{et al.}}: A Sample Article Using IEEEtran.cls for IEEE Journals}

\IEEEpubid{0000--0000/00\$00.00~\copyright~2021 IEEE}
\maketitle

\begin{abstract}
Recent years have witnessed huge successes in 3D object detection to recognize common objects for autonomous driving (e.g., \textit{vehicles} and \textit{pedestrians}). However, most methods rely heavily on a large amount of well-labeled training data. This limits their capability of detecting rare fine-grained objects (e.g., \textit{police cars} and \textit{ambulances}), which is important for special cases, such as emergency rescue, and so on. To achieve simultaneous detection for both common and rare objects, we propose a novel task, called generalized few-shot 3D object detection, where we have a large amount of training data for common (base) objects, but only a few data for rare (novel) classes. Specifically, we analyze in-depth differences between images and point clouds, and then present a practical principle for the few-shot setting in the 3D LiDAR dataset. To solve this task, we propose a simple and effective detection framework, including (1) an incremental fine-tuning method to extend existing 3D detection models to recognize both common and rare objects, and (2) a sample adaptive balance loss to alleviate the issue of long-tailed data distribution in autonomous driving scenarios. 
On the nuScenes dataset, we conduct sufficient experiments to demonstrate that our approach can successfully detect the rare (novel) classes that contain only a few training data, while also maintaining the detection accuracy of common objects.
\end{abstract}

\begin{IEEEkeywords}
Point cloud, few-shot, long-tail, autonomous vehicles, deep learning, 3D detection
\end{IEEEkeywords}

\begin{figure}[htbp] 
	\centering
	\includegraphics[width=0.48\textwidth]{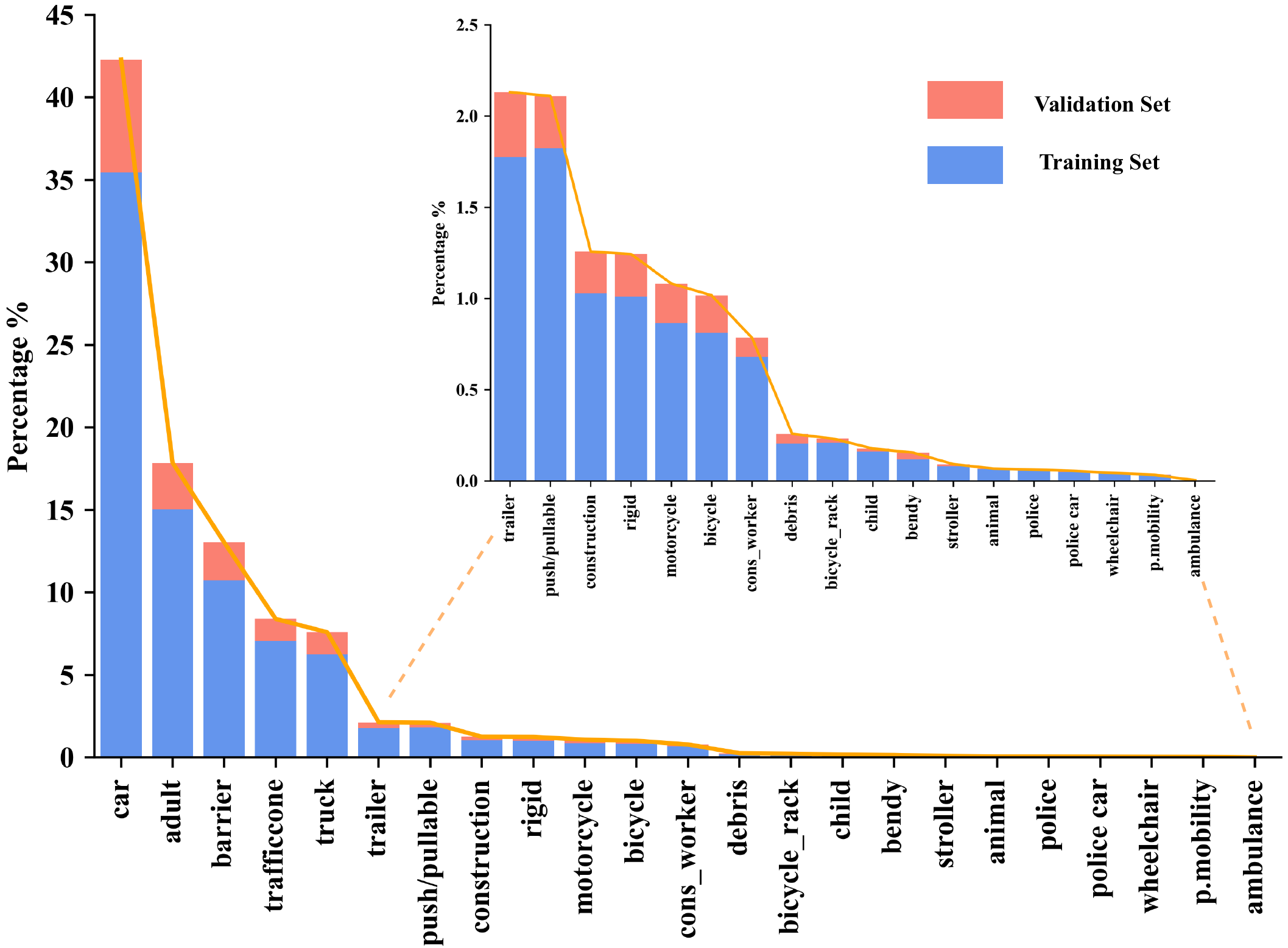} 
	\caption{Long tail problem in the nuScenes~\cite{caesar2020nuscenes} dataset.} 
	\label{fig:long tail} 
\end{figure}

\section{Introduction}
\IEEEPARstart{O}{bject} detection of the point cloud is a task of great interest, especially according to the flourishing of the application of 3D LiDAR in autonomous driving scenarios.
Many object detection works~\cite{li2021lidar, Mao_2021_ICCV} carry out the effort on a large amount of annotated LiDAR point cloud benchmarks, such as nuScenes~\cite{ caesar2020nuscenes}, KITTI~\cite{geiger2013vision}, Waymo~\cite{sun2020scalability}, ONCE~\cite{mao2021one}, and so on.
Most of them have a few categories of objects, such as Waymo and KITTI, where mainly three categories of objects, \textit{i.e.} \textit{vehicles}, \textit{pedestrians}, and \textit{bicycles}, are  considered for object detection. 
Many 3D works~\cite{shi2020pv, li2021lidar, cheng2021back, noh2021hvpr} have explored the detection of these three categories, and rely on extensive labeled data for learning.
It is worth noting that there are much more types of objects than these three in an autonomous driving scenario.
For example, nuScenes provides more detailed annotations of 23 types of objects.
However, most of the existing works cannot detect these uncommon objects, because of the shortage of samples.

\IEEEpubidadjcol

Detecting objects on the road with a wide range of categories is important and basic for environmental perception and autonomous driving.
For example, vehicles should avoid police cars or ambulances on the street, as they may carry out emergency pursuit or rescue work.
In addition, vehicles need to keep away from those objects of poor avoidance ability, such as strollers and wheelchairs, in order to ensure their safety.

Although these objects rarely appear on the street, once autonomous driving vehicles fail to detect them, they will not be able to make timely avoidance and cause traffic accidents.
So object detection for these uncommon categories is also of great interest. 
We have done a detailed statistical analysis of the nuScenes dataset, as shown in 
Fig.~\ref{fig:long tail}, there are 23 categories of annotated objects which represent a long-tail distribution in real-world scenarios.
Among the categories of objects on the street, a large number of objects are concentrated in a few categories, like \textit{car} and \textit{adult}.
Many categories, such as \textit{stroller} and \textit{police}, do not appear very often.
However, previous methods have chosen to intentionally ignore these uncommon categories,  because it is difficult to achieve better training results with less data.
In nuScenes dataset, only 10 large categories are detected and evaluated for actual detection. 
However, there are still good reasons to add more classes of objects to the original detection, even with a few training samples.
Inspired by 2D few-shot object detection, we would like to detect objects in these categories with a smaller number of samples.

In the 2D field, there exist many approaches to detect novel objects with only a few data available (few-shot detection) \cite{wang2020frustratingly, fan2021generalized, kukleva2021generalized, li2020overcoming} and alleviate the long-tail problem \cite{li2020overcoming, wang2021adaptive}. However, it is still a blank to detect rare objects with a few samples in terms of the 3D LiDAR point cloud. 
Directly bringing 2D few-shot solutions to 3D is infeasible.
Firstly, 3D LiDAR point clouds are very sparse compared to the images.
For example, most of the background areas in the whole scene are devoid of points, and an individual object may contain only a few points. The points will be concentrated on the surface of the object, where nearer points are denser and further ones are sparser. However, in image data, pixels are usually dense and uniform.
Moreover, the point cloud lacks the texture information in the image, which leads to difficulty in semantic analysis. For example, it is hard to distinguish between adults, police officers and construction workers using spatial structure information alone.
Thirdly, the number of objects contained in a point cloud scene is much larger than that in an image.
We meet more disturbances since there are many objects of the common classes and a very small number of the additional classes.
Therefore, to detect the additional classes of objects in the point cloud, it is not feasible to simply use the few-shot detection method in images for point clouds.

In this paper, we propose a practical principle for the few-shot setting in the 3D LiDAR dataset to detect additional classes of objects, by in-depth analyzing the differences between images and point clouds.
We propose a few-shot 3D object detection method. For uncommon category objects in the point cloud data, we merely require a few annotated objects in learning. Our method is trained at a low cost through training an existing 3D object detector. 
We should also note that although we want to focus on the uncommon category of objects, if the detection of common objects decreases significantly to detect objects that rarely appear, it is not worth the loss. So we aim to increase the detection of uncommon objects while the detection of common objects remains largely unchanged.
We utilize incremental learning for the additional categories, which ensures the model better generalizes to additional classes and better adapts to the long-tail distribution of the data in real life. 
In addition, to solve the category imbalance and hard sample disturbance during training, we design a sample adaptive balance (SAB) loss which focuses more on the distinction between objects of the additional and the common categories. 
Our method also reduces the mutual inhibitory effect between the common and the additional classes, thus retaining the detection rate of the common class and responding to objects of additional categories at the same time.

The main contributions of our work can be summarized as follows:
\begin{itemize}
	\item We are the first to explore the practical problem of few-shot 3D object detection of LiDAR point cloud, in order to detect additional categories of objects, which has an important meaning in autonomous driving.
	We put forward a task definition with a severe class imbalance problem.
	We provide a suitable dataset setting based on the nuScenes~\cite{caesar2020nuscenes} dataset for few-shot learning by studying different task splits for the multiple prediction heads of the network to learn.

\item We propose an incremental learning architecture to detect more categories. By adding each novel class to an individual branch, we avoid mutual interference between the common classes and the few-shot novel classes.
Our network has better generalization capabilities with the incremental branching setting.

\item We design a sample adaptive balance (SAB) loss function to deal with the severe long-tail distribution in point clouds. It balances the common and additional classes. Different from the Focal Loss, SAB loss reduces the disturbance of difficult negative samples. Our network learns the rare categories discriminatively and retains a reliable performance for base classes.

\item We conduct extensive experiments to explore the suitable setting of the dataset for incremental few-shot 3D object detection and test our method. The experimental results demonstrate that our model can detect additional objects of few-shot novel classes easily and effectively.
\end{itemize}

\section{Related Work}
This section presents the related works in two aspects:
3D object detection on LiDAR point cloud data and few-shot learning on 2D images.

\subsection{3D point cloud object detection}
3D point cloud object detection has been widely explored with a lot of data annotations ~\cite{li2021lidar, zhu2019class, Mao_2021_ICCV, Misra_2021_ICCV}. 
An intuitive idea is to transform the 3D point cloud into a 2D form from the bird-eye view (BEV)~\cite{lang2019pointpillars, zhang2020polarnet, zhu2020ssn, ge2020afdet} so that a 2D convolutional network could efficiently process on it. 
A better way to produce regular data organization from point clouds is to utilize a regular grid, such as VoxelNet~\cite{zhou2018voxelnet}.
Some works~\cite{yan2018second, tang2020searching} speed up the computation with sparse convolution. 
PointNet~\cite{qi2017pointnet} and its variants~\cite{qi2018frustum, qi2017pointnet++, wang2019dynamic, liu2019relation, wang2019pseudo} take the raw points as input, and solve the disordered  point cloud by  symmetric operations. 
PointRCNN~\cite{shi2019pointrcnn} firstly generates 3D region proposals to predict more accurate bounding box coordinates. 
PV-RCNN~\cite{shi2020pv} improves the performance by combining the voxel-based and the point-based method.
BtcDet~\cite{xu2022behind} alleviates the occlusion problem in the point cloud.
LiDAR RCNN~\cite{li2021lidar} gives a more precise prediction of the proposal.
CenterPoint~\cite{yin2021center} applies an anchor-free method.
Although these fully supervised methods have achieved good results, it is difficult to generalize to the few-shot novel classes in the autonomous driving scene.

\subsection{Few-shot learning}
Few-shot learning refers to learning a new class from few-shot annotated samples.
Meta-learning approaches~\cite{kang2019few, sun2019meta, li2021beyond, jamal2019task} train a meta-model on episodes of individual tasks. 
~\cite{qi2018frustum} simply fine-tunes the last layer of the detector on few-shot classes through a two-stage transfer learning. 
Following the idea of fine-tuning,~\cite{wu2020multi, sun2021fsce, kukleva2021generalized, qiao2021defrcn, fan2020few} achieve further improvements.
Few-shot 3D dataset also has a severe class imbalance problem inherent in 2D, but very little work has been conducted on 3D data. ~\cite{mondal2018few} and \cite{nie20203d} focus on few-shot learning on 3D medical image segmentation and 3D model classification. 
\cite{zhao2021few} starts to explore the few-shot problem on point cloud semantic segmentation tasks. But few-shot 3D object detection on the LiDAR point cloud has not yet been explored. 	
If we train novel classes step-wisely on a baseline, the networks suffer from catastrophic forgetting of the base classes \cite{li2018pami}. 
\cite{kukleva2021arxiv} designs a loss function to constrain the variant of network parameters and jointly fine-tune novel classes and base classes. \cite{shmelkov2017iccv} exploits two detection networks and a distillation loss. ~\cite{dong2021few} and \cite{tao2020few} utilize the graph topology-based knowledge distillation to transfer knowledge of old classes to the new classifier.
Some works treat few-shot learning by learning the novel classes incrementally ~\cite{wang2020frustratingly, fan2021generalized}.
Since it is important to perceive the few-shot novel classes and keep a reliable detection rate in autonomous driving, we inherit the core idea of incremental learning to deal with few-shot 3D object detection and proposed individual branches for each novel class to decrease the affection between the novel and the base classes.


\section{Generalized Few-Shot 3D Object Detection}
{Current mainstream methods of 3D detection in autonomous driving scenarios only focus on detecting common objects, which are usually seen on the road or street in the real world. However, detecting rare objects, such as police cars or strollers, are very critical but less explored topic for the autonomous driving system. Most existing models relying on large amounts of training data are not suitable to detect rare objects, since it is costly and more difficult to collect enough data for rare objects than common ones. For example, The nuScenes dataset collects 15 hours of driving data, but only obtains 26 rare `police', while the number of common `car' reaches 27701. If we want to provide enough amount of `police' similar to `car', we may need to collect 15,000 hours of driving data and pay more human-cost for annotation. Therefore, current methods are not practical and economical approaches to detecting rare objects. In this paper, we aim to explore the solution of detecting both common and rare objects, which can make full use of 3D driving data with long-tail distribution, \textit{i.e.}, 
the large number of common objects and a few rare objects. To do this, we propose a new task, called the generalized few-shot 3D object detection, 
and provide a reasonable set of benchmarks according to the characteristic and long-tail distribution of 3D driving data. Besides, we compare our 3D task with few-shot 2D object detection to explain the differences between these two tasks.}

\subsection{Few-Shot Setting for 3D Detection}
Inspired by few-shot 2D object detection ~\cite{wang2020frustratingly} , we define the generalized few-shot task for 3D object detection of LiDAR point cloud in autonomous driving scenarios. Firstly, the categories with a large number of annotated instances are used as base classes $C_b$. And the rarely appeared categories which only have a few instances are novel classes $C_n$. The base classes and the few-shot novel classes are non-overlapping, as formula $C_b \cap C_n = \oslash$. Then, our task aims to detect both classes on base and novel sets. The detailed setting with the existing dataset is as follows.

\noindent\textbf{Base Class Set.} 
Considering current mainstream datasets of 3D object detection, only the nuScenes dataset has the most categories and is more suitable for the few-shot 3D object detection task. Specifically, the nuScenes dataset provides 23 categories with annotated bounding boxes. However, only 10 common categories are used for the detection task, where \textit{construction workers}, \textit{police officers}, \textit{child} and \textit{adults} are merged as \textit{Pedestrain}, as well as \textit{bendy buses} and \textit{rigid buses} are merged as \textit{Bus}. The other nine categories are ignored. To maintain consistency with common 3D object detection methods, we select base classes $C_b$ by using the original 10 detection categories from the detection task, all of which have a large amount of labeled data. We use the original train and validation sets in nuScenes as the corresponding sets of our base classes, respectively.
For the novel classes, we try to use the ignored nine categories, however, we find that some categories are not suitable for our task, because of the too small amount of instances, and large intra-class variations in terms of morphology and size. 

\noindent\textbf{Novel Class Set.} We perform the statistical analysis of all classes in nuScenes. Specifically, we make statistics in both the training and the validation sets, and record the number of annotated instances for each category in all 1000 scenes of the dataset, as shown in Tab.~\ref{tab:statistics}. We can find some issues as follows.
	
\noindent\textit{(1) A handful of instances.} Following the setup of the few-shot 2D detection, for each novel class, we aim to randomly select 10 instances to build the training set and regard the others as testing samples. To evaluate the model's ability in few-shot learning, the number of testing samples is usually larger than the training sample in each novel class. According to this principle, we remove the \textit{wheelchair} and \textit{ambulance} classes, since the number of instances in each class is less than 20, and we can not obtain enough testing samples.

\noindent{\textit{(2) Too few key points in an instance.} {In the automatic driving scene, novel objects are often far away from our LiDAR car and there are serious occlusions. In addition, the volume of some objects is very small, so the number of points contained in the sample is often very small. For example, the category \textit{pensonal\_mobility} is defined as small electric or self-propelled vehicle, e.g. skateboard, segway, or scooter, on which the person typically travels in an upright position. Driver and (if applicable) rider should be included in the bounding box along with the vehicle. Obviously, these objects contain a small percentage of points compared to the points of the people in the same box. Most of the points in the box are provided by pedestrians, and it is hard to distinguish the difference between people and \textit{pensonal\_mobility} by such few points provided by the small vehicle. So we remove the \textit{pensonal\_mobility} class.}
}

\noindent\textit{(3) Large intra-class variations.} For the rest of the categories, we find that some categories are not fine-grained enough, but are grouped by a rather abstract term. Furthermore, they are not even a collection of objects with the same shape and characteristics. After visualizing the data of these categories, we found that the category of \textit{animals} includes cats, rats, dogs, birds, etc., and \textit{debris} includes trash bags, temporary road signs and tree branches, etc. The objects in these categories have large morphological and size variations and do not even share some common properties in terms of structure. This obviously can not be handled by only using point cloud data without color and texture information. Thus, we also remove the \textit{animal} and \textit{debris} classes.

After above analysis, we are only able to select four classes including \textit{police vehicle}, \textit{stroller}, \textit{pushable\_pullable} and \textit{bicycle\_rack} as the novel classes in our few-shot 3D object detection task. Notice that the \textit{pushable\_pullable} category has different objects, such as wheelbarrows, garbage bins with wheels, or shopping carts, however, they have similar shape properties. Thus, we can use this category for our task.

We adjust the validation set in these selected classes because of scarce or imbalanced instances.  
\textcolor{black}{
First of all, due to the small number of samples in the validation set, the evaluation results for rare classes on the official validation set are not convincing. Therefore, it is necessary to supplement some samples from the training set to enrich the number of samples of rare objects. Specifically, we ensure that there are 10 instances of training data for each class of objects in the training set, and we randomly select some of the remaining objects to add to the validation set. It is worth mentioning that objects in the rare category are heavily occluded. For example, we can find from Table II that more than 50\% of the visible areas of the stroller in the official validation set are less than 40\%. Therefore, in the process of randomly selecting objects from the training set to the validation set, we try to select some objects with high visibility rather than the full complement. The occlusion degree of each category in the supplemented validation set can be seen in Table II. To some extent, we alleviate the excessive occlusion of some categories of objects.
}

\begin{table}\scriptsize
	\setlength\tabcolsep{9pt}
	\caption{Instance statistics in the nuScenes dataset. A serious category imbalance exists. In the detection task, nuScenes merges adult, child, police\_officer and construction\_worker as pedestrian, bus\_bendy and bus\_rigid are merged as bus class. Classes marked with * are ignored in the nuScenes dataset. We choose 4 classes from the ignored classes before for few-shot learning, which are marked in bold. 
	}
	\centering
	\begin{tabular}{  l | l | l | l }
		\hline
		\textbf{Class}          & \textbf{train\&val}  & \textbf{train}  & \textbf{val} \\ 
		\hline
		adult                   & 10690      & 8870    & 1820  \\ 
		child                   & 141        & 122     & 19  \\ 
		police\_officer         & 34         & 31      & 3  \\ 
		construction\_worker    & 542        & 457     & 85  \\ 
		car                     & 27701      & 23158   & 4543  \\ 
		motorcycle              & 748        & 607     & 141  \\ 
		bicycle                 & 735        & 574     & 161  \\ 
		bus.bendy               & 85         & 66      & 19  \\ 
		bus.rigid               & 572        & 459     & 113  \\ 
		truck                   & 4215       & 3497    & 718  \\ 
		construction            & 648        & 529     & 119  \\ 
		trailer                 & 1114       & 919     & 195  \\ 
		barrier                 & 8415       & 6848    & 1567  \\ 
		trafficcone             & 6591       & 5381    & 1210  \\ 
		\hline  
		ambulance*              & 2          & 1       & 1  \\ 
		wheelchair*             & 18         & 18      & 0  \\ 
		\hline 
		personal\_mobility*     & 24         & 20      & 4  \\ 
		\hline
		animal*                 & 52         & 49      & 3  \\ 
		debris*                 & 164        & 124     & 40  \\ 
		\hline  
		\textbf{police}*                 & 26         & 24      & 2  \\ 
		\textbf{stroller}*      & 63         & 55      & 8  \\ 
		\textbf{bicycle\_rack}* & 122        & 108     & 14  \\ 
		\textbf{pushable\_pullable}*     & 1684       & 1473    & 211  \\ 
		\hline
	\end{tabular}
	\label{tab:statistics}
\end{table}
    
    

\noindent\textbf{Shot in 3D Detection.}
In few-shot 2D detection, an image is regarded as one shot, since it has contained a complete object. However, in 3D point cloud data, an instance usually has multiple frames captured by the LiDAR detection vehicle at different times and distances. If we only use a frame as one shot, we cannot utilize all information of an object. Thus, we treat all moments of samples of an instance as one shot of the object.
For a few-shot training set $\cal{P}$, we aggregate the point cloud data for all frames  of an instance as one shot and denote as:
\begin{equation}
	{\cal P}_{n,k}^{T} = \{p_{n,k}^{1}, p_{n,k}^{2}, ..., p_{n,k}^{T}\},	
\end{equation} 
where $n$ represents the $n^{th}$ category, $k$ means the $k^{th}$ instance, $t$ indicated the number of all frames of this instance.
So the N-way K-shot problem can be denoted as:
\begin{equation}
	{\cal P} = {\{({\cal P}_{1,k}^{T})_{k=1}^K,..., ({\cal P}_{N,k}^{T})_{k=1}^K\}}.
\end{equation}

\noindent\textbf{$\text{N}$-way $\text{K}$-shot Setting.}
	In 2D conditions, there are usually a few objects in an image. 
	As for the $N$-way $K$-shot setting, given an image, the network is usually provided with merely one object label of the novel class ignoring the other categories. 
	Differently, in 3D object detection, dozens to hundreds of objects exist in a scene. 
    {If we continue to follow the setting of 2D few-shot object detection by only giving the label of one novel object in a sample, and ignoring the labels of other objects, the model will suppress the objects of other classes in the same frame as the background}. This means that the objects of the same class may be suppressed as the background at one time and motivated as the detection target at another time, which is obviously unreasonable. 
    Furthermore, unlike 2D image data, the 3D LiDAR point cloud data in the autonomous driving scenario faces serious category imbalance. We believe that the ideal number of uniform control samples for each category is difficult to simulate the long-tailed distribution of categories in the actual situation. The setting of 2D few-shot object detection, \textit{i.e.}, the objects of both common and uncommon categories appear with the same frequency, is only an ideal situation and lacks some practical meaning for the 3D task. For this reason, instead of controlling that we enter one novel category object at a time, we choose to control that the total number of novel objects is N-way-k-shot. Specifically, for each input frame, we provide labels for all objects to be detected in it. We also allow several novel objects of different classes to appear in the same frame, and we control the total number of each novel object to be only k-shot.
    This setting is proposed by considering the long-tail distribution of 3D data, \textit{i.e.}, large amount of base (common) objects and a handful of novel (uncommon) objects. Thus, we remove the restriction of base classes to make full use of a large number of base labels.
    
\noindent\textbf{Goal of Our Task.}
According to the application of autonomous driving that LiDAR data are collected on the road, and the most common categories, \emph{i.e.} \textit{vehicles} and \textit{pedestrians}, should be primarily concerned in the few-shot 3D object detection task for traffic safety. 
Therefore, a reliable detection rate without apparent performance degradation for the common base classes is the prerequisite for pursuing the detection of rare novel categories.
Our goal for few-shot 3D object detection is to lightly train the network based on a few-shot setting per novel class to successfully detect novel objects, and maintain the performance of the base class detection.

\begin{table}
	\scriptsize
	\setlength\tabcolsep{3pt}
	\caption{The number of frames in the validation set.}
	\centering
	\label{tab:2}
	\begin{tabular}{  c|c | c | c | c | c | c}
		\hline
		\multicolumn{2}{c|}{\textbf{Visibility}}          & \textbf{0-40\%}  & \textbf{41-60\%}  & \textbf{61-80\%} & \textbf{81-100\%} 
		& \textbf{total}
		\\
		\hline 
		\multirow{2}{*}{\textbf{stroller}}
		& official   & 71   & 11   & 14 & 35 & 131
		\\
		\cline{2-7}
		& ours &109 &24 &30 &93 & 256
		\\
		\hline 
		\multirow{2}{*}{\textbf{police}}
		& official   & 7   & 0   & 3 & 63 & 73
		\\
		\cline{2-7}
		& ours &22 &12 &25 &162 & 221
		\\ 
		\hline 
		\multirow{2}{*}{\textbf{pushable\_pullable}}
		& official   & 861   & 200   & 253 & 2018 & 3332
		\\
		\cline{2-7}
		& ours &178 &71 &140 &951 & 1340
		\\ 
		\hline 
		\multirow{2}{*}{\textbf{bicycle\_rack}}
		& official   & 51   & 31   & 38 & 154 & 274
		\\
		\cline{2-7}
		& ours &41 &33 &101 &463 & 638
		\\ 
		\hline
	\end{tabular}
	\label{tab:difficulty}
\end{table}

\subsection{Comparison of 2D and 3D Few-Shot Object Detection}
	We analyze the differences between the 2D and 3D few-shot object detection tasks and emphasis the challenges in the 3D task.

\begin{figure*} 
\centering
\includegraphics[width=1\textwidth]{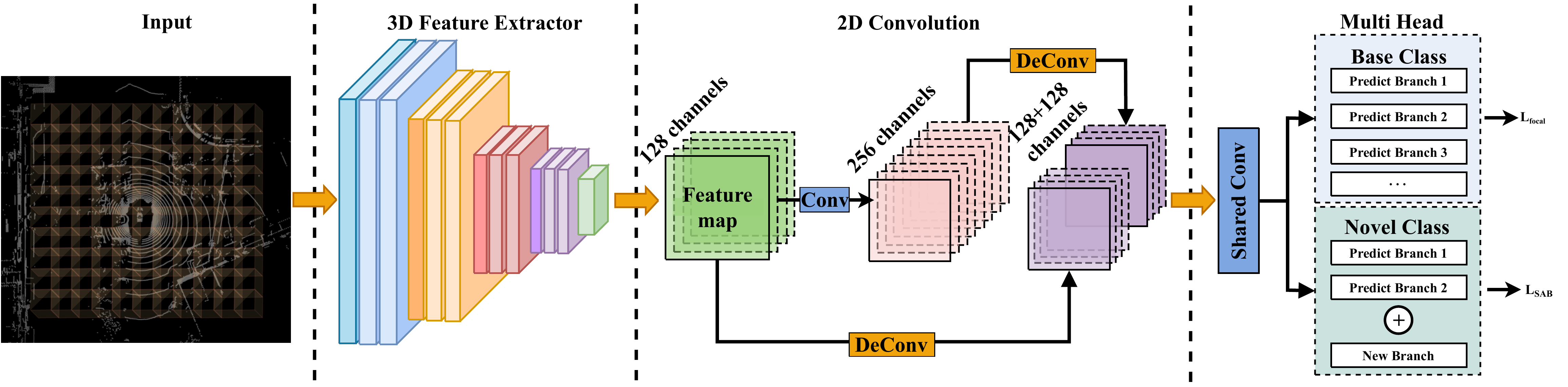} 
\caption{The framework of our method. An input 3D point cloud is processed by a 3D feature extractor and then fed into an RPN for further feature embeddings. After that, the features go through a shared convolutional layer, whose result will be sent to the multiple prediction heads for the final detection prediction. The few-shot 3D object detection adopts a fine-tuning manner. For an introduced few-shot novel class, the network adds an individual branch for it and updates merely the parameters of the new branch during learning.} 
\label{fig:framework} 
\end{figure*}

\noindent\textbf{Class imbalances in a scene.} 
    In a 2D few-shot object detection task, a picture usually contains only a few objects. 
    Thus, in a shot, there will be a similar number of base objects as the novel objects.
    However, in the 3D LiDAR point cloud scenario, on the one hand, a frame of a point cloud may contain hundreds of objects, while only one rare object may be included. This means that the number of objects in the base class and novel class in a frame is often more than 100:1, which is a serious class imbalance. On the other hand, in the face of point cloud input covering a range larger than 100m$\times$100m$\times$8m, there may be only one rare class object to be detected, and the size is only approximately 0.5m$\times$0.5m$\times$0.5m. In conclusion, for novel objects, the number is much less than that of base objects, and the space occupied by such novel objects is much smaller than that of common CAR objects. It is also likely to be submerged in a large number of other positive samples when processed by the model.
	Some previous solutions~\cite{zhu2019class, cai2021ace} deal with the imbalance problem by oversampling or data augmentation. 
	Since it is short of samples for novel classes, a large number of replications will bring about over-fitting.
	Thus these methods may not be suitable for few-shot 3D detection. 
	To overcome this problem, we design to introduce incremental learning to train the novel classes while keeping the discriminative ability for base classes. 
	Also, to improve our discriminability for the novel class, we design a dynamic loss function to improve the attention of the model for the novel class. The specific branch structure and design of the loss function will be described in detail in the next section.
	
\noindent\textbf{Quality of Samples.}
	In an image, the matrix of pixels constructs regular and dense input for 2D few-shot detection.
	It is perceived as a good instance that is unobstructed and of appropriate size. When an instance has some occlusions, it brings difficulty in detection. 
	However, it is worth noting that the unobscured areas of the instance are still represented with dense pixels that contain some semantics for inference.
	In the 3D scene, objects are formatted in a different way.
	The point cloud data is irregular and very sparse. 
	Besides, there are also a large number of occlusions of different degrees.
	The unoccluded regions of an object may merely contain a few points or even no points at all. 
	If there is a substantial distance between the object and the LiDAR vehicle, then the object would only contain a small number of points. 
	Moreover, the objects of the novel categories are not common on the road but are relatively far from the current vehicle with LiDAR equipment.
	This situation also brings more difficulty in detection, regardless of the object is being obscured or not.
	It is in line with common sense that, in an image, small objects are more difficult to detect compared with large objects, due to the deficiency of semantics provided by fewer pixels.
	These phenomenons reveal that the few-shot 3D object detection is more challenging than the 2D task.

\section{Method}

To evaluate the effectiveness of the normal 3D object detection methods on our new few-shot task, we select the SOTA model: CenterPoint~\cite{yin2021center}, as our baseline to detect base classes and add new incremental heads for the novel classes.

As shown in Fig.~\ref{fig:framework}, the framework consists of a voxelization of the point cloud, a 3D feature extractor, a 2D convolutional processing module and a multi-head module. After voxel the input point cloud, we get the 3D feature map by 3D convolution. Then the feature map is compressed into a 2D feature map from the Z-axis direction. The final feature map is predicted by the multi-head module to obtain the object class, center position, size and other information. 
In this section, we will introduce how to detect objects of novel classes and maintain the detection performance for the base class, by incremental finetuning and our elaborate loss function for this new problem.

\subsection{Fine-tuning Incremental Branches for Novel Classes}
Inspired by the 2D few-shot detection method, TFA~\cite{wang2020frustratingly}, 
we also adopt a two-stage fine-tuning approach for the detection network. 
In the first stage, namely base training, we continue to use the original base class training to obtain robust detection performance on base classes, by making full use of a large number of base class labels. In contrast to TFA, our goal of the second few-shot fine-tuning stage is to detect the novel class while maintaining the effectiveness of the base class. Thus, we choose to add one branch (head) for each novel class in the multi-head module to avoid disturbing the detection heads of base classes. 
Then we freeze the parameters of the baseline model and only finetune the incremental branches by using a few training data of novel classes.

\begin{figure} 
	\centering
	\includegraphics[width=0.5\textwidth]{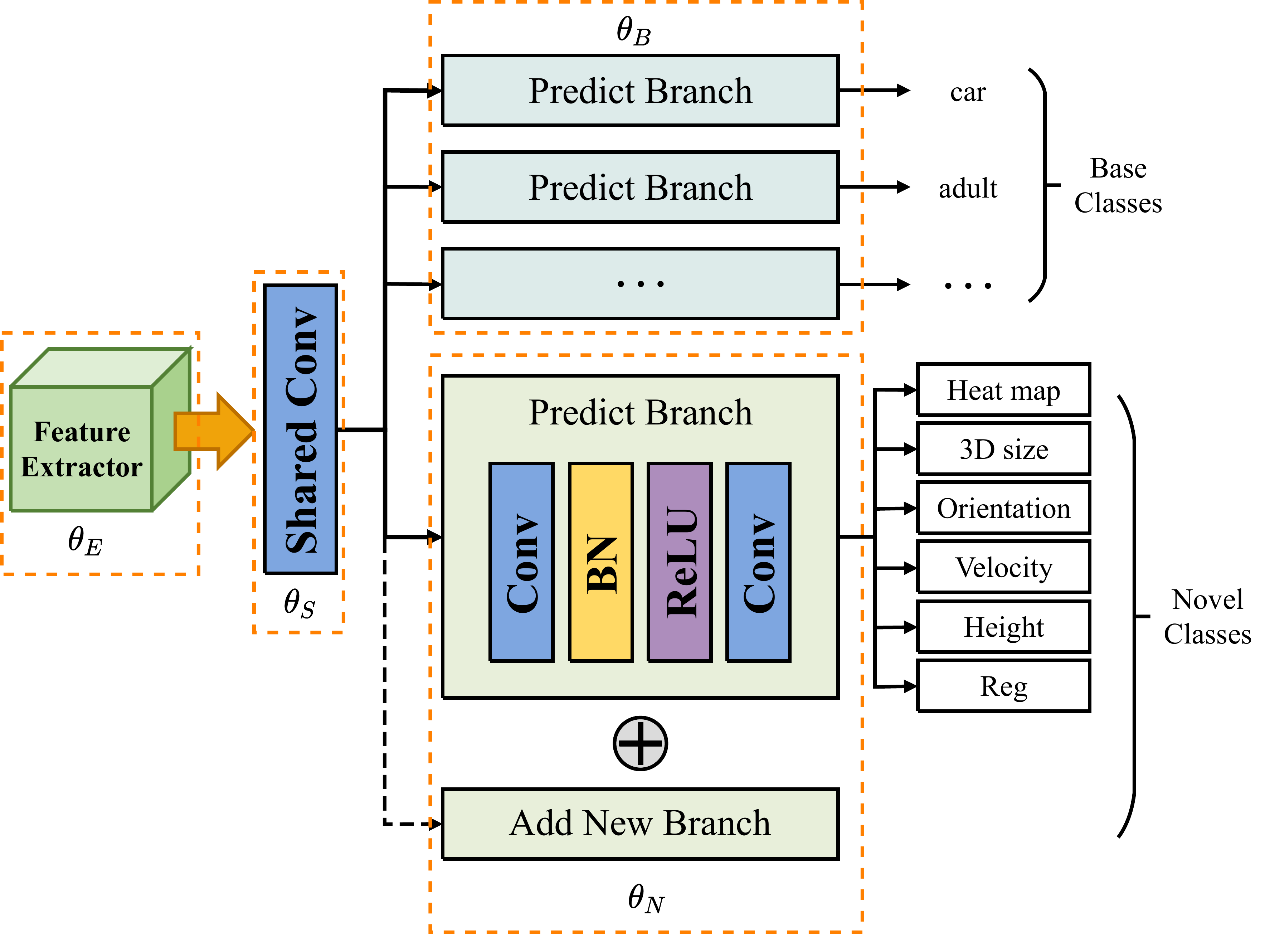}
	\caption{Incremental branches for few-shot novel classes. By adding a new individual branch for each novel class, we decrease the mutual affection between objects of the base and the novel classes.} 
	\label{fig:head} 
\end{figure}

Specifically, our network utilizes CenterPoint~\cite{yin2021center} as the baseline and VoxelNet~\cite{zhou2018voxelnet} as our backbone.
For few-shot learning, we retain the original 10 categories in the nuScenes dataset as the base classes.
We modify the head part of the network by adding new independent branches for each novel class, as shown in Fig. \ref{fig:framework}.
Each of the new branches has the same head as the base prediction branch of CenterPoint, which predicts the center position, size, orientation and velocity for the objects. Each incremental branch consists of a simple two-layer convolution, one layer of BN and one layer of ReLU. The details of the network structure can be seen in Fig. \ref{fig:head}.
After the first base-training stage, the backbone has obtained a good feature extraction ability in the first stage and the base head branches are discriminative for the base classes. Thus, in the second fine-tuning phase, we keep these parts of the network unchanged. We only train the incremental branches with our few-shot training set, to ensure the prediction for the base class does not decrease during the fine-tuning phase.

Our fine-tuning strategy is different from TFA, which fine-tunes all head branches for both base and novel classes. Specifically, TFA uses a small balanced training set consisting of the base and the novel classes, each of which is of K-shot. It freezes the parameters of the feature extractor and releases the parameters of the prediction heads, \textit{i.e.}, the final fully connected layers in the classification and regression parts for all classes. Notice that we only control the novel classes as $K$-shot in the input of fine-tuning phase, rather than using a small balanced training set similar to TFA. The goal is to investigate the effects of novel class detection in the case of class imbalance as much as possible.

Actually, we have tried to adopt the fine-tuning strategy of TFA, by simply fine-tuning all parameters in the last prediction part of the network, \textit{i.e}, the multi-head module. {However, in our experiments, we found that simply placing the base class and novel class in the same head would lead to catastrophic forgetting in base class detection. Because the base class and novel class objects may have similar shapes, causing the network to easily confuse the two classes of objects.}.
Besides, we also have tried different incremental branching methods and different training strategies in the second fine-tuning stage to explore a better incremental learning method for our few-shot 3D detection task. More details could be found in Sec. V.B. 

%

\subsection{Sample Adaptive Balance Loss}
During the actual fine-tuning process with a few samples, we find several challenges in the 3D scenario. 1) The scale and the number of objects in a novel class are too small compared with the large background area. This leads to a severe imbalance between the positive and the negative samples during fine-tuning. 2) During the first base-training stage, the network is apt to regard the objects of novel classes as the background if there are novel classes in the same scene. Thus, the novel classes are more inclined to be suppressed by the base-trained model. We need to correct these errors during fine-tuning.
3) Since the network is trained by many samples in base classes in the first stage, it usually generates high responses to the base classes. Thus, it is difficult to predict the novel classes by only using a few samples for fine-tuning, especially when the object of the base category is similar to the novel category.

To overcome the aforementioned challenges, we propose a novel sample adaptive balance loss, which is inspired by the focal loss \cite{lin2017focal}.
However, if we simply follow the original solution for 2D detection, there will be a relatively high weight of the foreground compared to the background. 
There may be a few novel class objects in the point cloud scenes while there are many objects of other classes around, and some negative samples with high response will occur during the detection process.
The small number of high-scored negative samples will be obliterated by a large number of low-scored negative samples.
Therefore, it is necessary to emphasize these difficult samples with a reasonable higher weight in 3D LiDAR point cloud scenes.
We design a new loss function to solve the conspicuous imbalance between positive and negative samples.

Inspired by the calculation of the weights in focal loss, we dynamically adjust the weights $w_{pos}$ of both the hard and the simple samples in the positive samples based on the confidence of the prediction $p$, which can be formulated as:
\begin{equation}
	w_{pos} = \sqrt{1-s},
\end{equation}
where $s$ is the score at the current position. 

The weight coefficient $w_{neg}$ is calculated based on the number of positive samples $num_{pos}$ and the number of negative samples $num_{neg}$ in each input sample to suppress the large number of background regions, which can be represented as:
\begin{equation}
	w_{neg} = \frac{num_{pos}}{num_{pos} + num_{neg}},
\end{equation}
In addition, we set a negative sample response threshold $\theta$ to measure the response level of the predicted regions, except for the area where the positive samples locate.
We statistic the number of the regions with responses higher than  $\theta$ particularly as $num_{hn}$.
Since these samples with higher confidence scores can easily be misclassified into other classes, we designed a dynamic weight $w_{hn}$ for these hard negative samples.
This weight will change dynamically during the training process depending on $num_{pos}$ and $num_{hn}$, as follows:
\begin{equation}
	w_{hn} = \frac{num_{hn}}{num_{hn} + num_{pos}}.
\end{equation}
Given a series of heatmaps $f_i$ with label $y_i$ from a sample, the total loss to learn can be calculated as:
\begin{equation}
	L_{\textit{\textit{SAB}}} = -\sum_{i=1}^{N} w_{i}[{y_i}log(f_i) + {(1-y_i)log(1-f_i)}],
\end{equation}
where $N$ denotes the total number of categories. For the object belonging to $k^{th}$ categories, $w_i$ indicates the proposed adaptive balance weight, which can be formulated as:
\begin{equation}
	w_{i} = \left\{
	\begin{aligned}
		w_{pos}, & \quad if \ i = k \\
		w_{neg}, & \quad if \ i \neq k \quad and \quad f_i \leq \theta \\
		w_{hn}, & \quad if \ i \neq k \quad and \quad f_i > \theta.
	\end{aligned}
	\right.
\end{equation}
The sample adaptive balance loss utilizes $w_{pos}$ to adjust the weights of hard and simple samples in positive samples, and uses $w_{neg}$ to reduce the impact of serious imbalance between the positive and the negative samples which is caused by a large number of background regions and finally exploits  $w_{hd}$ to pay more attention to hard negative samples with scores above the threshold. The setting of the threshold guarantees the generalization performance for the other categories. The proposed algorithm is detailed in Algorithm 1.

\begin{algorithm}[!thb]
\caption{SAB Loss For Each Training Sample}\label{alg:alg1}
\begin{algorithmic}
\STATE 
\STATE {\textsc{INPUT:}} $\textbf{Predict Output Map P},  \textbf{Ground Truth Map G}$
\STATE {\textsc{INITIALIZE:}} $\theta \gets 0.1, num_{pos} \gets 0, num_{neg} \gets 0, num_{hn} \gets 0$
\STATE $\textbf{for} \  y_{i,j} \in G \  \textbf{do}$
\STATE \hspace{0.5cm} $\textbf{if} \  { y_{i,j} = 1} \ \textbf{then}$
\STATE \hspace{1cm} $num_{pos} \gets num_{pos} + 1$
\STATE \hspace{0.5cm} $\textbf{else}$
\STATE \hspace{1cm} $num_{neg} \gets num_{neg} + 1$
\STATE \hspace{1cm} $\textbf{if} \  { f_{i,j} > \theta} \ \textbf{then}$
\STATE \hspace{1.5cm} $num_{hn} \gets num_{hn} + 1$
\STATE \hspace{1cm} $\textbf{end \ if}$
\STATE \hspace{0.5cm} $\textbf{end \ if}$
\STATE $\textbf{end \ for}$

\STATE $Update \ w_{pos} \  based \  on \  Equation(3)$
\STATE $Update \ w_{neg} \ based \ on \ Equation(4)$
\STATE $Update \ w_{hn} \ based \ on \ Equation(5)$

\STATE $\textbf{for} \  f_{i,j} \in Positive Sample Area \  \textbf{do}$
\STATE \hspace{0.5cm}$loss_{pos} \gets w_{pos} y_{i,j} \log(f_{i,j})$
\STATE $\textbf{end \ for}$
\STATE $\textbf{for} \  f_{i,j} \in Negative Sample Area \  \textbf{do}$
\STATE \hspace{0.5cm}$loss_{neg} \gets w_{neg} (1-y_{i,j})\log(1-f_{i,j})$
\STATE $\textbf{end \ for}$
\STATE $\textbf{for} \  f_{i,j} \in Hard Negative Sample Area \  \textbf{do}$
\STATE \hspace{0.5cm}$loss_{hn} \gets w_{hn} (1-y_{i,j})\log(1-f_{i,j})$
\STATE $\textbf{end \ for}$
\STATE $loss \gets loss_{pos}+loss_{neg}+loss_{hn}$
\STATE $\textbf{return} \ loss$

\end{algorithmic}
\label{alg1}
\end{algorithm}

We use the $L_\textit{SAB}$ on the heatmap to do the classification task, and use the $L_1$ loss as the loss function for regression tasks, including the prediction of size, orientation, and velocity.
Thus, the total loss can be expressed as:
\begin{equation}
	L = L_\textit{SAB} + \lambda L_{regression}
\end{equation}
To balance the classification loss and regression loss, we adjust their weights by $\lambda$, which is set to 0.25 in our experiments.

\begin{table*}\small
	\caption{Comparison of different fine-tuning strategies. Blue characters highlight the worst cases of catastrophic forgetting of the base classes. Bold characters indicate the best performances the model can achieve without catastrophic forgetting.}
		\setlength\tabcolsep{6pt}
	\centering	
		\begin{tabular}{ l|ccccc|ccccc|l}
		\hline
		\multirow{2}{*}{\textbf{Settings}}
		&\multicolumn{5}{c|}{\textbf{BASE}} 
		&\multicolumn{5}{c}{\textbf{NOVEL (10-shot)}} 
		&\multicolumn{1}{c}{\textbf{ALL}}
		\\
		\cline{2-12} 
		~
		& AP@0.5  & \textbf{$1.0$}  & \textbf{$2.0$} & \textbf{$4.0$} & \textbf{bmAP}
		& AP@0.5  & \textbf{$1.0$}  & \textbf{$2.0$} & \textbf{$4.0$} & \textbf{nmAP}
		& \textbf{cmAP}
		\\ 
		\hline 
		1. Baseline \textit{w.o.} fine-tuning
		& 65.3   & 73.7   & 76.4   & 77.8 &73.3
		& -   & -   & -   & - &-
		& -
		\\
		\hline 
		2. $\theta_E+\theta_S + \theta_B+\theta_N$
		& 42.3   & 51.6    & 54.8   & 56.7  & \textcolor{blue} {\textbf{51.4}}
		& 8.6   & 10.4    & 12.8   & 19.3  & 12.8
		& 40.4
		\\
		3. $\theta_E+\theta_S+\theta_N$
		& 11.1   & 13.2   & 14.2   & 15.4  & \textcolor{blue}{\textbf{13.4}}
		& 6.4   & 8.2   & 10.7   & 15.1  & 10.1
		& 12.5
		\\
		\hline
		4. $\theta_S+\theta_B+\theta_N$
		& 58.3   & 66.1   & 68.8   & 70.3  & 65.9
		& 3.4   & 4.1   & 5.9   & 8.4  & 5.4
		& 48.6
		\\
		5. $\theta_S+\theta_N$
		& 60.8   & 68.9   & 71.5   & 73.0  & 68.6  
		& 3.4   & 4.1   & 5.8   & 8.4  & 5.4
		& 50.5
		\\
		6. $\theta_B+\theta_N$
		& 58.9   & 66.9   & 69.2   & 70.8  & 66.5  
		& 3.4   & 3.8   & 5.3   & 7.6  & 5.0
		& 48.9
		\\
		7. $\theta_N$
		& \textbf{63.1}   & 71.1   & 73.8   & 75.2  & 70.8  
		& 3.9   & 4.3   & 6.3   & 9.0  & 5.9
		& 52.3
		\\
		\hline
		8. $\theta_B+\theta_N+L_{\textit{SAB}}$
		& 59.8 & 67.6 & 70.3 & 72.0 & 67.4 & 4.8 & 5.4 & 6.9 & 9.1 & 6.5
		& 50.0
		\\
	    9. $\theta_N+L_{\textit{SAB}}$
		& {63.0}   & \textbf{71.1}   & \textbf{73.8}   & \textbf{75.3}  & \textbf{70.8}  
		& \textbf{4.9}   & \textbf{5.4}   & \textbf{7.1}   & \textbf{9.5}  & \textbf{6.7}
		& \textbf{52.5}
		\\
		\hline 
	\end{tabular}
	\label{tab:finetune}
\end{table*}

\section{Experiments}
In this section, we conduct experiments on the nuScenes dataset. We first briefly describe the details of our training and evaluation setup. Then, we explore the effects of the model under strict novel class data quantity and quality.
In Section C, we explore the impact of each part of the network on the results. And in Section D, we further improve the detection results with our elaborate loss function. Next, we explore the impact of different head divisions on the detection of base and novel classes. In Section F, we perform ablation experiments about our method. Finally, we visualize the results on the nuScenes dataset.
\subsection{Implementation Details}
\textbf{Training details.} We utilize a two-stage training approach based on CenterPoint, with voxelnet as the backbone for all training stages.
As Fig.~\ref{fig:head} shows, the baseline is mainly composed of the feature extractor/RPN $\theta_E$, the shared convolutional layer $\theta_S$, and the base-class branches $\theta_B$.
The first stage of training is conducted on 10 base classes according to the setting of the original 3D object detection task.
We use a one-cycle learning rate variation with a maximum learning rate of 1e-3 and a minimum learning rate of 1e-4, for a total of 20 epochs. The batch size is set to 4, and the weight decay in the Adam optimizer is set to 0.01. 
When getting a reliable detector, the backbone of the detector is responsible for extracting underlying feature embeddings, and the last few layers are related to the high-level discriminative semantics for classification.
As a result, for the novel classes, we directly exploit the feature extraction part of the network and keep the parameters of $\theta_E$. 
We conduct experiments by adopting different training strategies on the baseline obtained in the first training stage.
In the second stage of training, the novel classes are trained on our constructed few-shot 3D point cloud training set.
For each novel category, we add a head branch to predict objects in that category.
In the second stage of training, the learning rate is set to one-tenth of the first stage, with training  80 epochs. 
We augment the novel class with GT-AUG~\cite{zhu2019class} strategy. That is, if there are less than two objects of a novel class in each input, we will copy some sample data and add them to the input. 
The objects of all categories are filtered according to the number of points, whose minimum number is set to 5 per object.
For the threshold value in $L_{\textit{SAB}}$, we set it to 0.1.
We design a group of training methods, as demonstrated in Tab.~\ref{tab:finetune}.

\textbf{Evaluating Details.}
To explore the detection accuracy of all classes, we calculate the average precision according to the official test metrics of nuScenes, which are based on the distance thresholds $\{0.5, 1, 2, 4\}$ meters between the predicted center and the ground truth location of the object. 
We keep the original settings of predicting at most 500 boxes for a whole point cloud scene. 
Setting 1 denotes the baseline without fine-tuning. 
We evaluate the original network on our validation set to acquire the detection results of the base class when new classes are not introduced.
In this way, we can further compare the degree of forgetting of the base class by this network after the introduction of novel class detection.
The current detection metrics only consider 10 significant classes in the detection task.
We adapt the evaluation metric of base classes and novel classes independently and set the detection range for each novel class.
While keeping the detection distance for the base class the same, we established different detection ranges for different classes of objects: 50m for police vehicles, 40m for strollers, and 30m for the other two novel classes.
Our evaluation criteria are divided into three parts. One is the mean average precision(mAP) of all base class objects, denoted as bmAP, to measure how well we detect common classes. One is the mAP of all novel objects, denoted as nmAP, to show how well we are detecting the novel objects. The last one is the mAP of all categories, which is used to represent the average detection effect, denoted as cmAP.

\subsection{Data quality of novel class objects}
Compared with the high-quality data of common objects, we first evaluate the data quality of novel objects. And we explore the detection ability of CenterPoint model for novel objects under the existing data conditions without considering the detection effect of Base objects.
Setting 2 represents the parameters of the whole network that are released to update in the second stage.
This setting is to explore the current performance of the network for detecting novel class objects, which are more difficult to detect, without considering the catastrophic forgetting of the base class. It is equivalent to finding the upper limit of our detection performance for the novel class without accounting for the loss of detection effectiveness of the base class.
In setting 3, we freeze the base-class branches and update the other components of the network.
Training that only considers the detection of novel class objects leads to the extraction of features in the feature extraction section that are more biased towards novel class objects and suppress base class objects. 

As we pursue a fast generalization to a few training samples, the base class will have a serious performance degradation due to the additional new classes of objects, which is unwanted in the autonomous driving scenario. 
It can be noticed that settings 2 and 3 bring a significant decrease in predicting the base class, as the bmAP metrics drop from 73.3\% to 51.4\% and 13.4\%, respectively (blue characters). Even if they obtain higher nmAP scores, they cannot be adopted in real-life scenarios.
At the same time, we can also see that the best result achieved by the existing model for the novel class without considering the base is only 12.8\%, which can be approximated as the upper limit of detection we can achieve in the case of poor quality of the novel class data.
Besides, it reveals that the feature extractor/RPN structure is not suitable to fine-tune in learning few-shot novel classes, since they are apt to provide satisfying features to guarantee the prediction accuracy for novel classes while forgetting the base classes. 

From the above experiments, it is not difficult to find that the base class objects have catastrophic forgetting. In order to ensure the detection effect of the base class, we try different training strategies to pursue that the detection effect of the base class is basically not decreased.

\subsection{Solving Catastrophic Forgetting}
In setting 4, only the feature extractor $\theta_E$ is frozen. 
By keeping the original feature extraction part of the model, the detection effect of the base class and novel class is considered only from the part of feature processing and classification.
Setting 5 indicated the shared convolutional layer $\theta_S$ and the novel-class branch will be updated.
We keep the branches of the base class completely and only train the parameters of the last layer of the shared feature and the novel class branches.
Setting 6 updates the base-class branches $\theta_B$ and the novel-class branches $\theta_N$.
In order to further preserve the detection effect of the base class, we do not change any step of the whole model for base class detection, but only add the novel branch to it and train the head part of the base class and novel class, and control the gradient back-propagation of the novel class branch will not affect the features processed by the base class.
Setting 7 merely updates the novel branches.
The purpose of this is to preserve the detection effect of the base class to the extreme so that the second stage of training is only for the novel branch.

From setting 4 and 5, the experiments show that by fixing the network parameters of $\theta_E$, we do effectively preserve the detection effect of the base class. 
This shows that keeping the feature extraction part of the model unchanged can effectively retain the very good feature extraction capability of the original model.
From setting 6 and 7 we found that keeping the shared convolution part of all branches before going to the head of each class preserves the feature extraction capability of the original network more effectively. Since the feature extraction part of the model is not changed, the effect of our base class is further preserved.
Furthermore, we discover that merely updating the novel-class branches $\theta_N$ achieves the best performance on both the base and the novel classes (bmAP reaches 70.8\% and nmAP is 5.9\%).
It is worth studying that in setting 7, we neither modified the head of the base class nor controlled the feature extraction part as before. Theoretically, the detection effect of the base class should be similar to the original model, but from the results, we found that 0.708 still has a 3\% decrease compared to 0.733. 
In order to investigate the reason for this part, we analyzed the experimental results and found that in the original case, the model would classify some of the background and base class objects that contain fewer points as base class objects, but now some novel class objects also appear in the classification options that contain fewer points, and the interference of these cases leads to a slight decrease in the detection effect of the base class.

\subsection{Alleviating sample imbalance}
The above settings investigate how to improve the detection of novel objects while preserving the effect of base class detection from the branching structure and training methods, and only consider reducing the effect of novel class detection on base class detection. In the background region and the base class object, some regions are easily confused with the novel class object, and these regions can interfere with the detection of the novel class object. So we propose sample adaptive balanced loss $L_\textit{SAP}$ to solve this problem. To verify the effectiveness of our loss function, we change the classification loss of the novel branch from focal loss to our $L_\textit{SAP}$ loss as setting 8 and setting 9 in setting 6 and setting 7.

In a point cloud scene, the number of these interference-filled regions is small compared to the number of positive samples of base class objects, but they are more compared to the number of novel class objects, which is relatively small. Therefore, in order to reduce the effect of these interference regions on the novel class detection, we propose the $L_\textit{SAP}$ loss and add the loss in settings 8 and 9. They can improve our result by 0.08 to 1.5 percentage points compared with settings 6 and 7.
By an additional SAP loss $L_\textit{SAP}$ (setting 9), the overall performance of the network is further improved. With an acceptable performance on the base classes (bmAP reaches 70.8\%), the nmAP metric reaches the best score of 6.7\%, which is 52.3\% of the 12.8\%.
From the average detection accuracy of each class, our method also achieves the best cmAP results.


\begin{figure*} 
	\centering
	\includegraphics[width=1\textwidth]{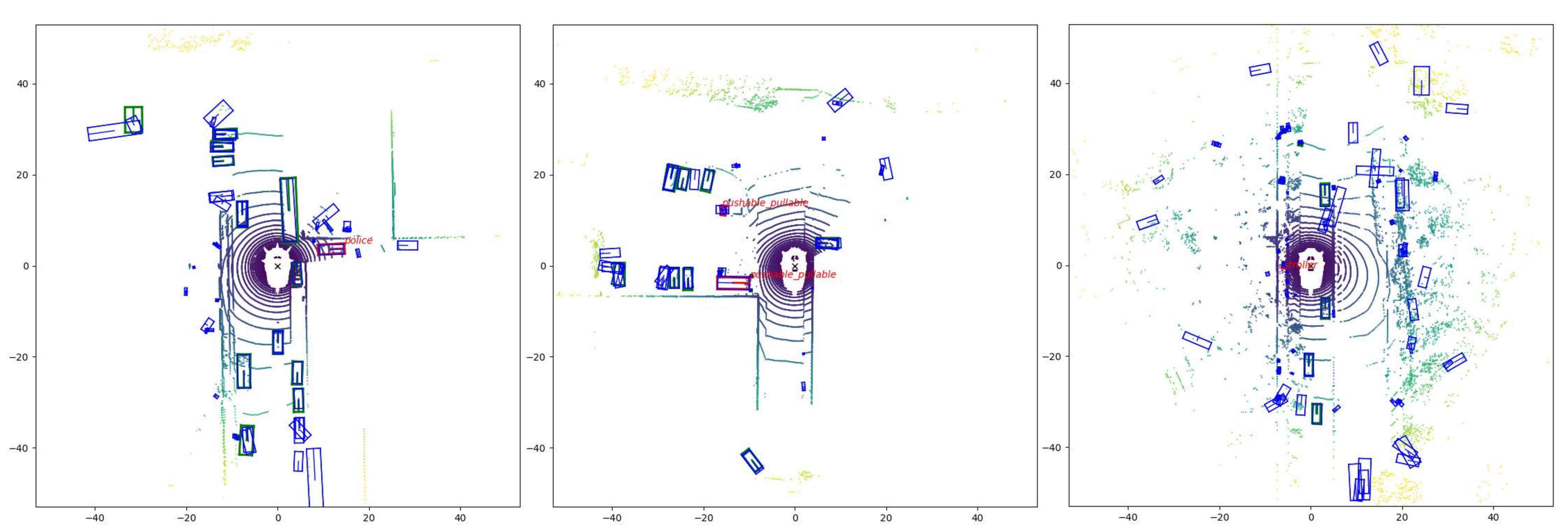}
	\caption{Visualization of the detection results from BEV. 
		The green boxes denote the ground truth of base classes, while the red indicates that of novel classes. The blue boxes are the prediction of our method. The network has a response to objects of novel classes via learning in incremental individual branches for few-shot novel categories. } 
	\label{fig:visualization} 
\end{figure*}

\begin{table}[t]\scriptsize
	\caption{Different split strategies and their performances. The different classes predicted by one head branch are listed in one brace. Novel classes are in bold.}
	\centering
	\setlength\tabcolsep{3pt}
	{
		\begin{tabular}{c | c | c | c}
			\hline
			\textbf{Split}  &	\textbf{Task} & \textbf{bmAP} & \textbf{nmAP}
			\\
			
			\hline 
			\multirow{4}{*}{\textbf{\#1}}  
			&\{car\}; \{truck, construction\_vehicle, \textbf{police}\};
			& \multirow{4}{*}{69.1} 
			& \multirow{4}{*}{2.1} \\
			&\{bus, trailer\};\{barrier\};\{motorcycle, bicycle\}; & &\\
			&\{pedestrian, \textbf{stroller}, traffic\_cone, \textbf{pushable\_pullable}\}; & &\\
			&\{\textbf{bicycle\_rack}\}; & &\\
			
			\hline 
			\multirow{4}{*}{\textbf{\#2}}    	
			&\{car\};\{truck, construction\_vehicle\};\{bus, trailer\};
			& \multirow{4}{*}{51.5} 
			& \multirow{4}{*}{2.6}  \\
			&\{barrier\};\{motorcycle, bicycle\};& &\\
			& \{pedestrian, traffic\_cone\}; & & \\
			&\{\textbf{stroller},\textbf{ pushable\_pullable},\textbf{ police}, \textbf{bicycle\_rack}\} & &\\

			\hline 
			\multirow{5}{*}{\textbf{\#3}}    	
			&\{car\} ;\{truck, construction\_vehicle\};\{bus, trailer\};
			& \multirow{5}{*}{70.8} 
			& \multirow{5}{*}{5.9}  \\
			&\{barrier\};\{motorcycle, bicycle\}; & &\\
			&\{pedestrian, traffic\_cone\};\{\textbf{stroller}\};  & &\\
			&\{\textbf{pushable\_pullable}\};\{\textbf{police}\};\{\textbf{bicycle\_rack}\} & &\\
			\hline 
	\end{tabular}}
	\label{tab:split}
\end{table}
\subsection{Different Split Settings}
The head part of the network categorized all classes into separate detection branches as individual detection tasks in real-world detection procedures. For the novel classes, we also investigated the impact of different categorization methods on the detection results.
In conventional 3D object detection on the point cloud, the principles lie in the size, shape and number of the object. 
The original 10 base classes in nuScenes are divided into different tasks.
Accordingly, we first analyze the novel classes and add them to the original tasks based on the size, shape, and sample size of each category (Tab.~\ref{tab:split} \#1).
Moreover, we do not limit adding to or changing the commonly used splits, but also try to train all novel classes in a new task, by utilizing a unified branch (Tab.~\ref{tab:split} \#2). 
In order to pursue a more straightforward extension to new categories of objects, we refer to some settings of incremental learning and treat each category as a separate task branch to predict (Tab.~\ref{tab:split} \#3).
The experimental results are shown in Tab.~\ref{tab:split}. 
As we can see, the incremental branch is best suitable for the task set.

\subsection{Analysis for the Loss Function}
To suppress high responses of particular background areas, we use SAB loss to supervise the learning of the novel classes. 
Experimental results (Tab.~\ref{tab:finetune}) show that the proposed SAB loss can better handle the sample imbalance. 
At the same time, we conduct experiments for a better setting of the hyperparameter $\theta$ in the $L_\textit{SAP}$ loss (Tab.~\ref{tab:loss}).
We exploit locations with confidence greater than a threshold of 0.1 as candidate regions for object centroids during inference. 
When the threshold value is below 0.1, the foreground object is suppressed and the detection is poor.
With different thresholds in the $L_\textit{SAP}$ loss, the bmAP and nmAP metrics change accordingly. We choose $\theta=0.1$ for it achieves the best nmAP score of 6.5\%. The results show that the $L_\textit{SAP}$ loss can suppress the difficult negative samples in the background region effectively.

\begin{table}\small
	\setlength\tabcolsep{5.5pt}
	\caption{hyper-parameters of $\theta$}
	\centering
	\begin{tabular}{  c | c | c | c | c | c}
		\hline
		& \textbf{$\theta=0.1$}  & \textbf{$\theta=0.2$}  & \textbf{$\theta=0.3$} & \textbf{$\theta=0.4$} & \textbf{$\theta=0.5$} \\
		\hline 
		\textbf{bmAP}& 70.8   & 70.8  & 70.8 & 70.8 & 70.8
		\\
		\textbf{nmAP} & \textbf{6.7}   & 5.4   & 6.2   & 5.0 & 5.3
		\\
		\hline
	\end{tabular}
	\label{tab:loss}
\end{table}
\subsection{Visualization} 
We also provided the visualization results in Fig.~\ref{fig:visualization}. The few-shot novel objects represent small scales. Overall, we produce favorable prediction boxes for basically all objects in the whole point cloud. Even in a complex scene environment, we can detect novel objects from a large number of base class objects.

\section{Conclusion}
{Our work is the first attempt to explore few-shot 3D object detection of LiDAR point cloud data for autonomous driving scenarios to detect additional classes.} To address the issue that real-world datasets are often long-tail distributed, we achieve object detection for new data categories by simply training on incremental branches with only a small number of samples. We also propose a new loss, namely the SAB loss, to improve the detection performance of novel classes. Although our work achieved significant improvements in detecting 3D point clouds with little training data, there is still a long way to go for few-shot 3D object detection of the point cloud. Many challenging problems are still unsolved which we will further discuss in our future work.

\bibliographystyle{IEEEtran}
\bibliography{IEEEabrv, egbib}

\begin{thebibliography}{10}
\providecommand{\url}[1]{#1}
\csname url@samestyle\endcsname
\providecommand{\newblock}{\relax}
\providecommand{\bibinfo}[2]{#2}
\providecommand{\BIBentrySTDinterwordspacing}{\spaceskip=0pt\relax}
\providecommand{\BIBentryALTinterwordstretchfactor}{4}
\providecommand{\BIBentryALTinterwordspacing}{\spaceskip=\fontdimen2\font plus
\BIBentryALTinterwordstretchfactor\fontdimen3\font minus
  \fontdimen4\font\relax}
\providecommand{\BIBforeignlanguage}[2]{{%
\expandafter\ifx\csname l@#1\endcsname\relax
\typeout{** WARNING: IEEEtran.bst: No hyphenation pattern has been}%
\typeout{** loaded for the language `#1'. Using the pattern for}%
\typeout{** the default language instead.}%
\else
\language=\csname l@#1\endcsname
\fi
#2}}
\providecommand{\BIBdecl}{\relax}
\BIBdecl

\bibitem{caesar2020nuscenes}
H.~Caesar, V.~Bankiti, A.~H. Lang, S.~Vora, V.~E. Liong, Q.~Xu, A.~Krishnan,
  Y.~Pan, G.~Baldan, and O.~Beijbom, ``nuscenes: A multimodal dataset for
  autonomous driving,'' in \emph{Proceedings of the IEEE/CVF conference on
  computer vision and pattern recognition}, 2020, pp. 11\,621--11\,631.

\bibitem{li2021lidar}
Z.~Li, F.~Wang, and N.~Wang, ``Lidar r-cnn: An efficient and universal 3d
  object detector,'' in \emph{Proceedings of the IEEE/CVF Conference on
  Computer Vision and Pattern Recognition}, 2021, pp. 7546--7555.

\bibitem{Mao_2021_ICCV}
J.~Mao, Y.~Xue, M.~Niu, H.~Bai, J.~Feng, X.~Liang, H.~Xu, and C.~Xu, ``Voxel
  transformer for 3d object detection,'' in \emph{Proceedings of the IEEE/CVF
  International Conference on Computer Vision (ICCV)}, October 2021, pp.
  3164--3173.

\bibitem{geiger2013vision}
A.~Geiger, P.~Lenz, C.~Stiller, and R.~Urtasun, ``Vision meets robotics: The
  kitti dataset,'' \emph{The International Journal of Robotics Research},
  vol.~32, no.~11, pp. 1231--1237, 2013.

\bibitem{sun2020scalability}
P.~Sun, H.~Kretzschmar, X.~Dotiwalla, A.~Chouard, V.~Patnaik, P.~Tsui, J.~Guo,
  Y.~Zhou, Y.~Chai, B.~Caine \emph{et~al.}, ``Scalability in perception for
  autonomous driving: Waymo open dataset,'' in \emph{Proceedings of the
  IEEE/CVF Conference on Computer Vision and Pattern Recognition}, 2020, pp.
  2446--2454.

\bibitem{mao2021one}
J.~Mao, M.~Niu, C.~Jiang, H.~Liang, J.~Chen, X.~Liang, Y.~Li, C.~Ye, W.~Zhang,
  Z.~Li \emph{et~al.}, ``One million scenes for autonomous driving: Once
  dataset,'' \emph{arXiv preprint arXiv:2106.11037}, 2021.

\bibitem{shi2020pv}
S.~Shi, C.~Guo, L.~Jiang, Z.~Wang, J.~Shi, X.~Wang, and H.~Li, ``Pv-rcnn:
  Point-voxel feature set abstraction for 3d object detection,'' in
  \emph{Proceedings of the IEEE/CVF Conference on Computer Vision and Pattern
  Recognition}, 2020, pp. 10\,529--10\,538.

\bibitem{cheng2021back}
B.~Cheng, L.~Sheng, S.~Shi, M.~Yang, and D.~Xu, ``Back-tracing representative
  points for voting-based 3d object detection in point clouds,'' in
  \emph{Proceedings of the IEEE/CVF Conference on Computer Vision and Pattern
  Recognition}, 2021, pp. 8963--8972.

\bibitem{noh2021hvpr}
J.~Noh, S.~Lee, and B.~Ham, ``Hvpr: Hybrid voxel-point representation for
  single-stage 3d object detection,'' in \emph{Proceedings of the IEEE/CVF
  Conference on Computer Vision and Pattern Recognition}, 2021, pp.
  14\,605--14\,614.

\bibitem{wang2020frustratingly}
X.~Wang, T.~E. Huang, T.~Darrell, J.~E. Gonzalez, and F.~Yu, ``Frustratingly
  simple few-shot object detection,'' \emph{arXiv preprint arXiv:2003.06957},
  2020.

\bibitem{fan2021generalized}
Z.~Fan, Y.~Ma, Z.~Li, and J.~Sun, ``Generalized few-shot object detection
  without forgetting,'' in \emph{Proceedings of the IEEE/CVF Conference on
  Computer Vision and Pattern Recognition}, 2021, pp. 4527--4536.

\bibitem{kukleva2021generalized}
A.~Kukleva, H.~Kuehne, and B.~Schiele, ``Generalized and incremental few-shot
  learning by explicit learning and calibration without forgetting,'' in
  \emph{Proceedings of the IEEE/CVF International Conference on Computer
  Vision}, 2021, pp. 9020--9029.

\bibitem{li2020overcoming}
Y.~Li, T.~Wang, B.~Kang, S.~Tang, C.~Wang, J.~Li, and J.~Feng, ``Overcoming
  classifier imbalance for long-tail object detection with balanced group
  softmax,'' in \emph{Proceedings of the IEEE/CVF conference on computer vision
  and pattern recognition}, 2020, pp. 10\,991--11\,000.

\bibitem{wang2021adaptive}
T.~Wang, Y.~Zhu, C.~Zhao, W.~Zeng, J.~Wang, and M.~Tang, ``Adaptive class
  suppression loss for long-tail object detection,'' in \emph{Proceedings of
  the IEEE/CVF Conference on Computer Vision and Pattern Recognition}, 2021,
  pp. 3103--3112.

\bibitem{zhu2019class}
B.~Zhu, Z.~Jiang, X.~Zhou, Z.~Li, and G.~Yu, ``Class-balanced grouping and
  sampling for point cloud 3d object detection,'' \emph{arXiv preprint
  arXiv:1908.09492}, 2019.

\bibitem{Misra_2021_ICCV}
I.~Misra, R.~Girdhar, and A.~Joulin, ``An end-to-end transformer model for 3d
  object detection,'' in \emph{Proceedings of the IEEE/CVF International
  Conference on Computer Vision (ICCV)}, October 2021, pp. 2906--2917.

\bibitem{lang2019pointpillars}
A.~H. Lang, S.~Vora, H.~Caesar, L.~Zhou, J.~Yang, and O.~Beijbom,
  ``Pointpillars: Fast encoders for object detection from point clouds,'' in
  \emph{Proceedings of the IEEE/CVF Conference on Computer Vision and Pattern
  Recognition}, 2019, pp. 12\,697--12\,705.

\bibitem{zhang2020polarnet}
Y.~Zhang, Z.~Zhou, P.~David, X.~Yue, Z.~Xi, B.~Gong, and H.~Foroosh,
  ``Polarnet: An improved grid representation for online lidar point clouds
  semantic segmentation,'' in \emph{Proceedings of the IEEE/CVF Conference on
  Computer Vision and Pattern Recognition}, 2020, pp. 9601--9610.

\bibitem{zhu2020ssn}
X.~Zhu, Y.~Ma, T.~Wang, Y.~Xu, J.~Shi, and D.~Lin, ``Ssn: Shape signature
  networks for multi-class object detection from point clouds,'' in
  \emph{Computer Vision--ECCV 2020: 16th European Conference, Glasgow, UK,
  August 23--28, 2020, Proceedings, Part XXV 16}.\hskip 1em plus 0.5em minus
  0.4em\relax Springer, 2020, pp. 581--597.

\bibitem{ge2020afdet}
R.~Ge, Z.~Ding, Y.~Hu, Y.~Wang, S.~Chen, L.~Huang, and Y.~Li, ``Afdet: Anchor
  free one stage 3d object detection,'' \emph{arXiv preprint arXiv:2006.12671},
  2020.

\bibitem{zhou2018voxelnet}
Y.~Zhou and O.~Tuzel, ``Voxelnet: End-to-end learning for point cloud based 3d
  object detection,'' in \emph{Proceedings of the IEEE conference on computer
  vision and pattern recognition}, 2018, pp. 4490--4499.

\bibitem{yan2018second}
Y.~Yan, Y.~Mao, and B.~Li, ``Second: Sparsely embedded convolutional
  detection,'' \emph{Sensors}, vol.~18, no.~10, p. 3337, 2018.

\bibitem{tang2020searching}
H.~Tang, Z.~Liu, S.~Zhao, Y.~Lin, J.~Lin, H.~Wang, and S.~Han, ``Searching
  efficient 3d architectures with sparse point-voxel convolution,'' in
  \emph{European Conference on Computer Vision}.\hskip 1em plus 0.5em minus
  0.4em\relax Springer, 2020, pp. 685--702.

\bibitem{qi2017pointnet}
C.~R. Qi, H.~Su, K.~Mo, and L.~J. Guibas, ``Pointnet: Deep learning on point
  sets for 3d classification and segmentation,'' in \emph{Proceedings of the
  IEEE conference on computer vision and pattern recognition}, 2017, pp.
  652--660.

\bibitem{qi2018frustum}
C.~R. Qi, W.~Liu, C.~Wu, H.~Su, and L.~J. Guibas, ``Frustum pointnets for 3d
  object detection from rgb-d data,'' in \emph{Proceedings of the IEEE
  conference on computer vision and pattern recognition}, 2018, pp. 918--927.

\bibitem{qi2017pointnet++}
C.~R. Qi, L.~Yi, H.~Su, and L.~J. Guibas, ``Pointnet++: Deep hierarchical
  feature learning on point sets in a metric space,'' \emph{arXiv preprint
  arXiv:1706.02413}, 2017.

\bibitem{wang2019dynamic}
Y.~Wang, Y.~Sun, Z.~Liu, S.~E. Sarma, M.~M. Bronstein, and J.~M. Solomon,
  ``Dynamic graph cnn for learning on point clouds,'' \emph{Acm Transactions On
  Graphics (tog)}, vol.~38, no.~5, pp. 1--12, 2019.

\bibitem{liu2019relation}
Y.~Liu, B.~Fan, S.~Xiang, and C.~Pan, ``Relation-shape convolutional neural
  network for point cloud analysis,'' in \emph{Proceedings of the IEEE/CVF
  Conference on Computer Vision and Pattern Recognition}, 2019, pp. 8895--8904.

\bibitem{wang2019pseudo}
Y.~Wang, W.-L. Chao, D.~Garg, B.~Hariharan, M.~Campbell, and K.~Q. Weinberger,
  ``Pseudo-lidar from visual depth estimation: Bridging the gap in 3d object
  detection for autonomous driving,'' in \emph{Proceedings of the IEEE/CVF
  Conference on Computer Vision and Pattern Recognition}, 2019, pp. 8445--8453.

\bibitem{shi2019pointrcnn}
S.~Shi, X.~Wang, and H.~Li, ``Pointrcnn: 3d object proposal generation and
  detection from point cloud,'' in \emph{Proceedings of the IEEE/CVF conference
  on computer vision and pattern recognition}, 2019, pp. 770--779.

\bibitem{xu2022behind}
Q.~Xu, Y.~Zhong, and U.~Neumann, ``Behind the curtain: Learning occluded shapes
  for 3d object detection,'' in \emph{Proceedings of the AAAI Conference on
  Artificial Intelligence}, vol.~36, no.~3, 2022, pp. 2893--2901.

\bibitem{yin2021center}
T.~Yin, X.~Zhou, and P.~Krahenbuhl, ``Center-based 3d object detection and
  tracking,'' in \emph{Proceedings of the IEEE/CVF Conference on Computer
  Vision and Pattern Recognition}, 2021, pp. 11\,784--11\,793.

\bibitem{kang2019few}
B.~Kang, Z.~Liu, X.~Wang, F.~Yu, J.~Feng, and T.~Darrell, ``Few-shot object
  detection via feature reweighting,'' in \emph{Proceedings of the IEEE/CVF
  International Conference on Computer Vision}, 2019, pp. 8420--8429.

\bibitem{sun2019meta}
Q.~Sun, Y.~Liu, T.-S. Chua, and B.~Schiele, ``Meta-transfer learning for
  few-shot learning,'' in \emph{Proceedings of the IEEE/CVF Conference on
  Computer Vision and Pattern Recognition}, 2019, pp. 403--412.

\bibitem{li2021beyond}
B.~Li, B.~Yang, C.~Liu, F.~Liu, R.~Ji, and Q.~Ye, ``Beyond max-margin: Class
  margin equilibrium for few-shot object detection,'' in \emph{Proceedings of
  the IEEE/CVF Conference on Computer Vision and Pattern Recognition}, 2021,
  pp. 7363--7372.

\bibitem{jamal2019task}
M.~A. Jamal and G.-J. Qi, ``Task agnostic meta-learning for few-shot
  learning,'' in \emph{Proceedings of the IEEE/CVF Conference on Computer
  Vision and Pattern Recognition}, 2019, pp. 11\,719--11\,727.

\bibitem{wu2020multi}
J.~Wu, S.~Liu, D.~Huang, and Y.~Wang, ``Multi-scale positive sample refinement
  for few-shot object detection,'' in \emph{European Conference on Computer
  Vision}.\hskip 1em plus 0.5em minus 0.4em\relax Springer, 2020, pp. 456--472.

\bibitem{sun2021fsce}
B.~Sun, B.~Li, S.~Cai, Y.~Yuan, and C.~Zhang, ``Fsce: Few-shot object detection
  via contrastive proposal encoding,'' in \emph{Proceedings of the IEEE/CVF
  Conference on Computer Vision and Pattern Recognition}, 2021, pp. 7352--7362.

\bibitem{qiao2021defrcn}
L.~Qiao, Y.~Zhao, Z.~Li, X.~Qiu, J.~Wu, and C.~Zhang, ``Defrcn: Decoupled
  faster r-cnn for few-shot object detection,'' in \emph{Proceedings of the
  IEEE/CVF International Conference on Computer Vision}, 2021, pp. 8681--8690.

\bibitem{fan2020few}
Q.~Fan, W.~Zhuo, C.-K. Tang, and Y.-W. Tai, ``Few-shot object detection with
  attention-rpn and multi-relation detector,'' in \emph{Proceedings of the
  IEEE/CVF Conference on Computer Vision and Pattern Recognition}, 2020, pp.
  4013--4022.

\bibitem{mondal2018few}
A.~K. Mondal, J.~Dolz, and C.~Desrosiers, ``Few-shot 3d multi-modal medical
  image segmentation using generative adversarial learning,'' \emph{arXiv
  preprint arXiv:1810.12241}, 2018.

\bibitem{nie20203d}
J.~Nie, N.~Xu, M.~Zhou, G.~Yan, and Z.~Wei, ``3d model classification based on
  few-shot learning,'' \emph{Neurocomputing}, vol. 398, pp. 539--546, 2020.

\bibitem{zhao2021few}
N.~Zhao, T.-S. Chua, and G.~H. Lee, ``Few-shot 3d point cloud semantic
  segmentation,'' in \emph{Proceedings of the IEEE/CVF Conference on Computer
  Vision and Pattern Recognition}, 2021, pp. 8873--8882.

\bibitem{li2018pami}
Z.~Li and D.~Hoiem, ``Learning without forgetting,'' in \emph{IEEE Transactions
  on Pattern Analysis and Machine Intelligence}, 2018, pp. 2935--2947.

\bibitem{kukleva2021arxiv}
A.~Kukleva, H.~Kuehne, and B.~Schiele, ``Generalized and incremental few-shot
  learning by explicit learning and calibration without forgetting,'' 2021.

\bibitem{shmelkov2017iccv}
K.~Shmelkov, C.~Schmid, and K.~Alahari, ``Incremental learning of object
  detectors without catastrophic forgetting,'' in \emph{2017 IEEE International
  Conference on Computer Vision (ICCV)}, 2017, pp. 3420--3429.

\bibitem{dong2021few}
S.~Dong, X.~Hong, X.~Tao, X.~Chang, X.~Wei, and Y.~Gong, ``Few-shot
  class-incremental learning via relation knowledge distillation,'' in
  \emph{Proceedings of the AAAI Conference on Artificial Intelligence},
  vol.~35, no.~2, 2021, pp. 1255--1263.

\bibitem{tao2020few}
X.~Tao, X.~Hong, X.~Chang, S.~Dong, X.~Wei, and Y.~Gong, ``Few-shot
  class-incremental learning,'' in \emph{Proceedings of the IEEE/CVF Conference
  on Computer Vision and Pattern Recognition}, 2020, pp. 12\,183--12\,192.

\bibitem{cai2021ace}
J.~Cai, Y.~Wang, and J.-N. Hwang, ``Ace: Ally complementary experts for solving
  long-tailed recognition in one-shot,'' in \emph{Proceedings of the IEEE/CVF
  International Conference on Computer Vision}, 2021, pp. 112--121.

\bibitem{lin2017focal}
T.-Y. Lin, P.~Goyal, R.~Girshick, K.~He, and P.~Doll{\'a}r, ``Focal loss for
  dense object detection,'' in \emph{Proceedings of the IEEE international
  conference on computer vision}, 2017, pp. 2980--2988.

\end{thebibliography}
\vfill

\end{document}